%% file: sample-sigconf.tex
\documentclass[sigconf, nonacm=true]{acmart}
\AtBeginDocument{%
  }

\copyrightyear{2025}
\acmYear{2025}
\setcopyright{rightsretained}
\acmConference[GECCO '25]{Genetic and Evolutionary Computation Conference}{July 14--18, 2025}{Malaga, Spain}
\acmBooktitle{Genetic and Evolutionary Computation Conference (GECCO '25), July 14--18, 2025, Malaga, Spain}
\acmDOI{}
\acmISBN{}

\input{Parts/Notations_Imports}

\begin{document}

\title{
\Longframework{}: A Generic Approach for Quality-Diversity in Noisy, Stochastic or Uncertain Domains
}

\author{Manon Flageat}
\email{manon.flageat18@imperial.ac.uk}
\affiliation{
  \institution{AIRL, Imperial College London}
  \city{London}
  \country{UK}
}

\author{Johann Huber} 
\email{johann.huber@isir.upmc.fr}
\affiliation{
  \institution{ISIR, Sorbonne Universite}
  \city{Paris}
  \country{France}
}

\author{François Helenon} 
\email{francois.helenon@isir.upmc.fr}
\affiliation{
  \institution{ISIR, Sorbonne Universite}
  \city{Paris}
  \country{France}
}

\author{Stephane Doncieux}
\email{stephane.doncieux@upmc.fr}
\affiliation{%
  \institution{ISIR, Sorbonne Universite}
  \city{Paris}
  \country{France}
}

\author{Antoine Cully}
\email{a.cully@imperial.ac.uk}
\affiliation{
  \institution{AIRL, Imperial College}
  \city{London}
  \country{UK}
}

\renewcommand{\shortauthors}{Flageat et al.}

\begin{abstract}
  \input{Parts/0_Abstract}
\end{abstract}




\maketitle

\input{Parts/1_Introduction}

\input{Parts/2_Background}

\input{Parts/3_Methods}

\input{Parts/4_Experiments}

\balance
\input{Parts/5_Conclusion}

\begin{acks}
    \input{Parts/Acknowledgement}
\end{acks}

\bibliographystyle{ACM-Reference-Format}
\bibliography{Parts/Biblio}

\newpage
\appendix

\input{Parts/Appendix}

\end{document}

%% file: Parts/Notations_Imports.tex
\newcommand{\Longframework}[0]{Extract-QD Framework}
\newcommand{\framework}[0]{EQD Framework}

\newcommand{\Longlongname}[0]{Extract-MAP-Elites}
\newcommand{\Longname}[0]{Extract-ME}
\newcommand{\name}[0]{EME}

\newcommand{\Longnamepga}[0]{Extract-PGA}
\newcommand{\namepga}[0]{EPGA}

\usepackage{booktabs} 
\usepackage{makecell}
\usepackage{enumitem}
\setlist[enumerate]{leftmargin=4.5mm, label=\alph*)}
\setlist[itemize]{leftmargin=4.5mm}

%% file: Parts/0_Abstract.tex
Quality-Diversity (QD) has demonstrated potential in discovering collections of diverse solutions to optimisation problems. Originally designed for deterministic environments, QD has been extended to noisy, stochastic, or uncertain domains through various Uncertain-QD (UQD) methods.
However, the large number of UQD methods, each with unique constraints, makes selecting the most suitable one challenging.
To remedy this situation, we present two contributions: first, the \Longframework{} (\framework{}), and second, \Longname{} (\name{}), a new method derived from it.
The \framework{} unifies existing approaches within a modular view, and facilitates developing novel methods by interchanging modules. 
We use it to derive \name{}, a novel method that consistently outperforms or matches the best existing methods on standard benchmarks, while previous methods show varying performance.
In a second experiment, we show how our \framework{} can be used to augment existing QD algorithms and in particular the well-established Policy-Gradient-Assisted-MAP-Elites method, and demonstrate improved performance in uncertain domains at no additional evaluation cost. 
For any new uncertain task, our contributions now provide \name{} as a reliable "first guess" method, and the \framework{} as a tool for developing task-specific approaches.
Together, these contributions aim to lower the cost of adopting UQD insights in QD applications.

%% file: Parts/1_Introduction.tex
\section{Introduction}

\begin{figure}[t!]
  \centering
  \includegraphics[width=\linewidth]{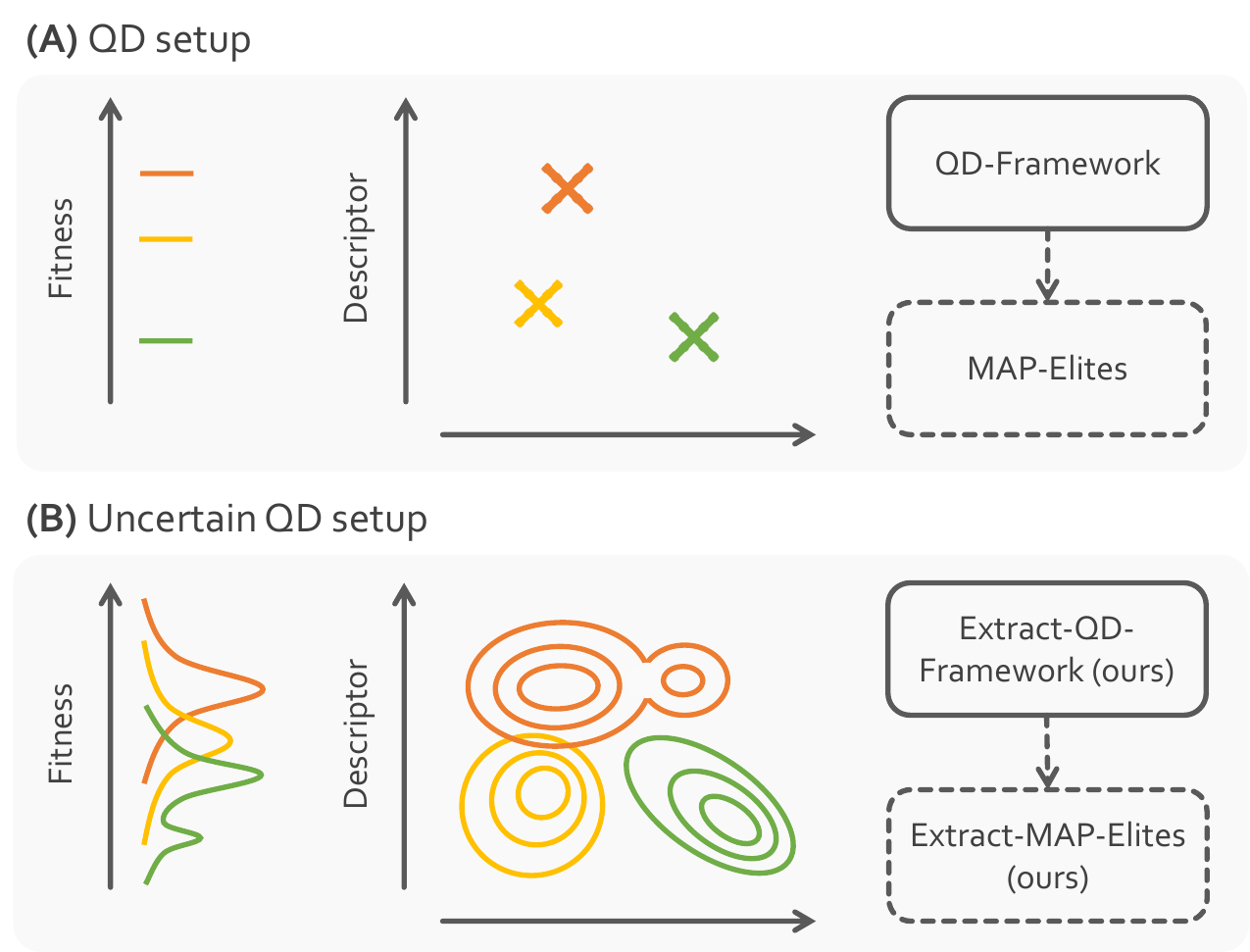}
  \caption{
    Comparison of the (A) standard QD setup, where each solution gets a single deterministic fitness and descriptor feature value, and the (B) UQD setup, where each solution gets a distribution of these values. 
    We propose the \Longframework{} and \Longlongname{} for UQD, equivalent respectively to the QD-Framework and MAP-Elites for QD. 
  }
  \Description{Motivation for our contributions}
  \label{fig:uqd}
\end{figure}


Problems in decision-making and content generation benefit from finding a diverse set of high-performing solutions. Such diversity enables users to choose among a variety of effective options~\cite{map_elites}, serves as a repository of potential solutions for unforeseen situations~\cite{nature}, or enables to identify stepping stones that might otherwise be overlooked~\cite{novelty_search}. 
Quality-Diversity (QD) algorithms~\cite{pugh, book_chapter, framework} have emerged as a powerful approach to uncovering such sets of diverse and high-performing solutions. 
To do so, they not only evaluate the solution's performance, known as its fitness but also characterise its way of solving the task (i.e. its behaviour), known as its feature descriptor, which allows them to quantify its novelty with respect to other solutions.
Thus, QD algorithms have found applications across various domains, including discovering diverse robotic controllers~\cite{nature, huber2023quality}, generating synthetic datasets for model training~\cite{huber2024qdgset, gaier2024generative} or adversarial examples for model testing~\cite{samvelyan2024rainbow}, and producing innovative designs~\cite{gaier2017generative} or video game content~\cite{gravina, earle2022illuminating}. 


Despite their potential, a key limitation of QD algorithms is their reliance on deterministic evaluations: they traditionally assume that a single evaluation of a solution reliably estimates its fitness and feature descriptor. 
However, this assumption does not hold true for many real-world tasks involving noise or stochasticity. 
For a motivating example, let's consider a QD method used to find a collection of controllers for a legged robot~\cite{nature}. If the sensor used to estimate the fitness (for example the walking speed) is noisy, the same controller may get different fitness from one evaluation to another. Thus, a slow controller could get "lucky" and appear the fastest thanks to a sensor glitch. QD elitism would lead this "lucky" to replace a truly good-performing one, resulting in a final population that contains underperforming controllers.
Similarly, if the sensor that estimates the feature descriptor incorrectly identifies a controller as walking efficiently on three legs when it actually uses all four, this "lucky" controller would replace genuine three-legged controllers, as none would match its speed. Consequently, the final population would fail to include true three-legged controllers, resulting in a drastic loss of diversity and potentially failing to meet user's requirements. 
Additionally, uncertainty in QD tasks extends beyond simple sensor noise to include stochastic environment dynamics~\cite{flageat2023uncertain, tjanaka2023training, mace2023quality, flageat2024exploring} and noise inherent to the task definition~\cite{huber2024domain, huber2024speeding}. This uncertainty also extends beyond robotics and has been observed in various QD tasks, such as content generation~\cite{earle2022illuminating} or design applications~\cite{gaier2024generative}.
This setup where the fitnesses and descriptor features of solutions are distributions instead of single deterministic values is known as Uncertain QD (UQD)~\cite{flageat2023uncertain} and illustrated in Figure~\ref{fig:uqd}.


Prior work has proposed multiple algorithms to tackle the challenges of UQD~\cite{mace2023quality, flageat2020fast, flageat2023uncertain, flageat2024exploring, grillotti2023don, adaptive}. 
While these algorithms are promising, they all have different constraints and limitations. 
For example, Adaptive-Sampling~\cite{adaptive} rely on sequential evaluations, making it slow in parallelisable contexts, while Deep-Grid~\cite{flageat2020fast} assumes specific noise structure, and Archive-Sampling~\cite{flageat2023uncertain} requires sufficiently small populations to allow for periodic re-evaluation of their entire content.
This complex landscape of UQD approaches makes it difficult to know which one to try when facing a new UQD task. 
In this work, we aim to provide users with a set of accessible tools to bring UQD insights into their own QD applications.


To this end, we present two main contributions: first, the \Longframework{} (\framework{}), and second, an instantiation of it, \Longname{} (\name{}).
The \framework{} proposes a new view on existing UQD approaches, allowing to encompass them all under the same common modular view. 
Thus, it also facilitates defining new UQD approaches by swapping modules.
\name{} is a new instantiation of the \framework{} that demonstrates strong performances across commonly studied tasks. 
Together, these two contributions aim to provide accessible tools to tackle any new uncertain QD application: \name{} acts as a solid "first guess" method, likely to perform well, while the \framework{} serves as a comprehensive toolbox for developing new tailored UQD algorithms.


To demonstrate the benefits of our contributions, we present two experimental studies.
First, we evaluate \name{} on a set of common UQD tasks from prior work. Our results show that \name{} consistently performs at least as well as the best-performing method for each task, unlike previous methods, which exhibit varying performance across tasks. Second, we explore how integrating our \framework{} with widely-used QD approaches leads to significant performance improvements. Here we consider the well-established PGA-MAP-Elites~\cite{nilsson2021policy} algorithm and show how accounting for uncertainty using our framework drastically improves its performance on uncertain tasks at no additional evaluation cost. 
We hope these contributions help lower the cost of integrating UQD insights into QD algorithms, thereby expanding their applicability.

%% file: Parts/2_Background.tex
\section{Background and Related Work}

\subsection{Quality-Diversity} \label{sec:qd}

Quality-Diversity (QD) optimisation~\cite{book_chapter, pugh, framework} aims to find collections of solutions to an optimisation task that are both diverse and high-performing. While traditional optimisation algorithms focus on producing a single high-fitness solution, QD algorithms instead generate an archive $\mathcal{A}$ of solutions: a population of solutions that maximise fitness while maintaining diversity across predefined dimensions. These diversity dimensions are typically defined as part of the optimisation task, we refer to them as feature descriptors.

\subsubsection{MAP-Elites} The most widely used QD method is \textbf{MAP-Elites (ME)}~\cite{map_elites}, illustrated in Figure~\ref{fig:algo}.A. ME partitions the feature descriptor space into a grid of cells and retains only the highest-performing solution in each cell, known as the elite of the cell. The collection of elites constitutes the archive $\mathcal{A}$. To populate and refine $\mathcal{A}$, ME operates iteratively over a fixed number of generations. In each generation, a batch of solutions is copied uniformly from $\mathcal{A}$ and mutated to produce offspring, which are then evaluated and added back into $\mathcal{A}$. 
In ME, new offspring are added to $\mathcal{A}$ if they either occupy an empty cell or outperform the current elite in their cell. 
The main alternative to ME is Novelty Search with Local Competition (NSLC)~\cite{lehman2011evolving}. 
All methods in this work build upon ME, following previous work in QD applied to uncertain domains, but our proposed \framework{} includes NSLC-based approaches. 

\subsubsection{QD Framework} \label{sec:qd_framework}
The QD Framework~\cite{framework} aims to unify QD algorithms, be they ME-based or NSLC-based. It proposes to decompose QD algorithms into two main building blocks: (1) the container, which gathers all the solutions found so far into an ordered collection (the grid in ME), and (2) the selection operator, which chooses the solutions used as parents to generate the next generation of offspring (uniform sampling from the grid in ME).
In this work, we propose an extension of this framework for uncertain tasks in QD. 


\subsection{Uncertain Quality-Diversity} \label{sec:uqd}

Standard QD algorithms assume that the fitness $f_i$ and feature descriptor $d_i$ of a solution $i$ can be reliably estimated from a single evaluation. 
On the contrary, Uncertain QD (UQD)~\cite{flageat2023uncertain} assumes that each solution can get a distribution of possible fitness and feature descriptor values: $f_i \sim \mathcal{D}_{f_i}$ and $d_i \sim \mathcal{D}_{d_i}$. This distinction is illustrated in Figure~\ref{fig:uqd}.
This Section covers existing UQD works, defining the sub-problems that arise in UQD setups and introducing existing approaches. 
We note that optimization under uncertainty has been independently extensively studied within the Evolutionary Computation (EC) community~\cite{ea_uncertain, ea_uncertain_2}.

\subsubsection{Uncertain Quality-Diversity Sub-Problems} \label{sec:problems}
Three main sub-problems emerge from the UQD literature: 
\begin{enumerate}
    \item \textbf{Performance Estimation:} the first problem in UQD is to accurately estimate the expected fitness and descriptor features of solutions within a trackable number of samples~\cite{adaptive, flageat2020fast, flageat2023uncertain}. 
    QD algorithms usually rely on a single evaluation of each solution. Due to their elitism, they tend to retain “lucky” solutions: solutions with overestimated $f$ or misplaced $d$, caused by favourable but unrepresentative (i.e. outlier) evaluations. 
    Thus, single-sample approaches do not translate well to UQD tasks. 
    On the other end of the spectrum, approaches that evaluate every solution enough times to obtain highly accurate estimates of $f$ and $d$ are likely to be computationally intractable.
    Thus, effectively distributing samples across solutions to accurately estimate solutions' performance and keep truly high-performing and diverse solutions in the archive $\mathcal{A}$ is a crucial issue in UQD. 
    
    \item \textbf{Reproducibility Maximisation:} the second problem in UQD is to prioritise reproducible solutions over non-reproducible ones with similar performance.
    Reproducibility~\cite{flageat2023uncertain, grillotti2023don, mace2023quality} refers to the ability of a solution to consistently reproduce its feature descriptor across multiple evaluations (i.e. tight spread of feature descriptor distribution). Reproducibility is a desirable property in QD as it provides guarantees to the final users that solutions will behave similarly when used in a downstream task~\cite{huber2024domain, mace2023quality}. 
    For example, when finding a set of solutions that can move to different target positions, it is desirable to get solutions that consistently reach their target position all or most of the time. 
    Thus, when given the choice between two solutions that perform equivalently but have different reproducibility, UQD algorithms should favour the most reproducible one. 
    It is important to note that this problem does not always arise as in certain tasks, all solutions have the same reproducibility. 
    
    \item \textbf{Performance-Reproducibility Trade-off:} the third UQD problem is to implement preferences among solutions that present different trade-offs between fitness and reproducibility~\cite{flageat2024exploring}.
    In many UQD tasks, fitness and reproducibility are not aligned: for instance, a controller that enables a robot to walk very quickly (high fitness) may also be more prone to lead the robot to fall (low reproducibility) compared to a slower controller (lower fitness but higher reproducibility). As a result, there may not be a clearly dominating solution for the task. 
    Thus, UQD algorithms should be able to enforce preferences specified by a user over this trade-off. 
    Again, this third UQD problem only arises in tasks where solutions get different reproducibility. 
\end{enumerate}
For simplicity and clarity, this paper primarily focuses on the Performance Estimation problem. However, we detail in the next sections all existing methods tackling all three problems, and we show that our \framework{} generalizes to all of them. 

\subsubsection{Fixed-Sampling UQD Approaches} \label{sec:fixed_sampling}
This is the most common type of UQD approach. They re-evaluate each solution $N$ times, where $N$ is a fixed number, to estimate its expected fitness and feature descriptor before adding it to the archive $\mathcal{A}$.
\textbf{ME-Sampling}~\cite{hbr}, the most common fixed-sampling approach, simply adds solutions to $\mathcal{A}$ based on the average of the $N$ reevaluations, thus only focusing on the Performance Estimation problem (see Section~\ref{sec:problems}).
\textbf{ME-Reprod}~\cite{grillotti2023don}, \textbf{ME-Low-Spread}~\cite{mace2023quality}, \textbf{ME-Weighted}~\cite{flageat2024exploring}, \textbf{ME-Delta}~\cite{flageat2024exploring} and \textbf{MOME-Reprod}~\cite{flageat2024exploring} are all variants of ME-Sampling that also tackle the other two sub-problems of UQD. To do so, they use the $N$ samples to also estimate solution reproducibility and optimise different combinations of the fitness and reproducibility values. 
Despite their ease of implementation, fixed-sampling approaches induce high sampling costs, limiting their applications. Additionally, they have been shown to hinder exploration by overlooking fragile solutions that might be key intermediate step toward promising solutions~\cite{flageat2023uncertain, flageat2024exploring, ea_uncertain_2}.

\subsubsection{Adaptive-Sampling UQD Approaches} \label{sec:adaptive_sampling}
This type of UQD approach, inspired by similar works in EC~\cite{ea_adaptive, ea_adaptive_2}, aims to lower the sampling cost of fixed-sampling approaches by distributing samples more wisely. To do so, instead of evaluating every solution a fixed number of times, they only reevaluate promising solutions~\cite{adaptive, flageat2023uncertain}. 
In \textbf{Adaptive-Sampling-ME (Adapt-ME)}~\cite{adaptive}, an offspring solution can replace an elite only if it is still better after having been sampled the same number of times as this elite. 
Additionally, each time a new solution proves less performing than an elite, this elite is re-sampled, leading powerful elites to be more and more evaluated, increasing the certainty about their values. As re-evaluated elites may prove to belong to another cell than the one they were initially evaluated into, elites might need to "drift" to other cells during the optimisation. To prevent a cell from suddenly becoming empty, Adapt-ME keeps multiple candidate elites per cell. 
This is known as a "depth", for example, an archive of depth $d=8$ would keep $8$ elites in each cell. Among these $8$ elites, only $1$, the best performing one, or "top" elite is returned in the final archive $\mathcal{A}$, but the others are used during the optimisation to replace drifting elites. 
While Adapt-ME has proven promising, its major drawback is that its evaluations can only be partially parallelisable, as the number of evaluations for one solution is directly dependent on the evaluations of other solutions.
This reason led to propose \textbf{Archive-Sampling (AS)}~\cite{flageat2023uncertain}, a simple variant of Adapt-ME that simply re-evaluates all the solutions in the archive $\mathcal{A}$ at each generation. AS also keeps a depth $d$ of elites, and re-evaluates the elites in the depth with the one in the top-layer at each generation. 
While simpler than Adapt-ME, AS has been shown to significantly outperform fixed-sampling approaches~\cite{flageat2023uncertain,flageat2024exploring}, while Adapt-ME has shown more contrasted results~\cite{adaptive}. To the best of our knowledge, these two approaches have not been directly compared before. 
While Adapt-ME and AS only explicitly tackle the Performance Estimation problem (see Section~\ref{sec:problems}), \textbf{AS-Weighted}~\cite{flageat2024exploring} and \textbf{AS-Delta}~\cite{flageat2024exploring} are both variants of AS that also address the other two sub-problems of UQD by approximating the reproducibility of solutions from re-evaluations.

\subsubsection{Other UQD Approaches}  \label{sec:deep_grid}
Two other approaches have been proposed for UQD that do not belong to the two previous categories. 
\textbf{Deep-Grid}~\cite{flageat2020fast} is based on the observation that "lucky" solutions (those with overestimated fitness due to stochastic evaluations) tend to dominate archive cells, preventing more representative solutions from replacing them. This can escalate over time, as only increasingly luckier solutions can displace these outliers, leading to unrealistic fitness values within the archive.
To counteract this effect, Deep-Grid constantly "questions" elites: it allows them to be replaced by any solution, even those that appear lower performing. To still optimise performance, Deep-Grid maintains a depth $d$ of elites (see adaptive-sampling approaches) and selects parents fitness-proportionally within cells.
Alternatively, \textbf{ARIA}~\cite{grillotti2023don} is an optimisation module that aims to improve the reproducibility of the solutions contained in the final population returned by any QD algorithm. While ARIA tackles the UQD setup, it has been designed to be run on top of another QD algorithm, and requires an order of magnitude more evaluations than any other UQD algorithm, making it hard to compare. 
For these reasons, we do not include it in our comparison. 
Finally, we note that several works have highlighted the benefits of gradient-based and evolution-strategy-based mutations in promoting reproducible solutions~\cite{colas2020scaling, flageat2023empirical, flageat2024enhancing, mace2023quality, faldor2023synergizing, tjanaka2023training}. 

%% file: Parts/3_Methods.tex
\section{Method}

This section introduces our two contributions: the \Longframework{} (\framework{}) and one of its instantiations \Longname{} (\name{}). 
When faced with a new UQD task, \name{} constitutes a solid "first guess" method, while our \framework{} is meant as a toolbox to develop tailored UQD algorithms. 
For clarity, we introduce \name{} first, before generalising it into the \framework{}.

\subsection{\Longname{}}

\name{} is an adaptive-sampling UQD approach (see Section~\ref{sec:adaptive_sampling}) that tackles uncertainty by re-evaluating part of the elites in the archive $\mathcal{A}$ during the optimisation process. 
As shown in Figure~\ref{fig:algo}, \name{} follows the same loop as ME but adds three components: an extraction mechanism, an estimation mechanism and an archive depth. 
In the following, we introduce each of these three components and detail the relation between \name{} and existing similar approaches. 

\subsubsection{Extraction} \label{sec:extraction}
\begin{figure}[t!]
  \centering
  \includegraphics[width=\linewidth]{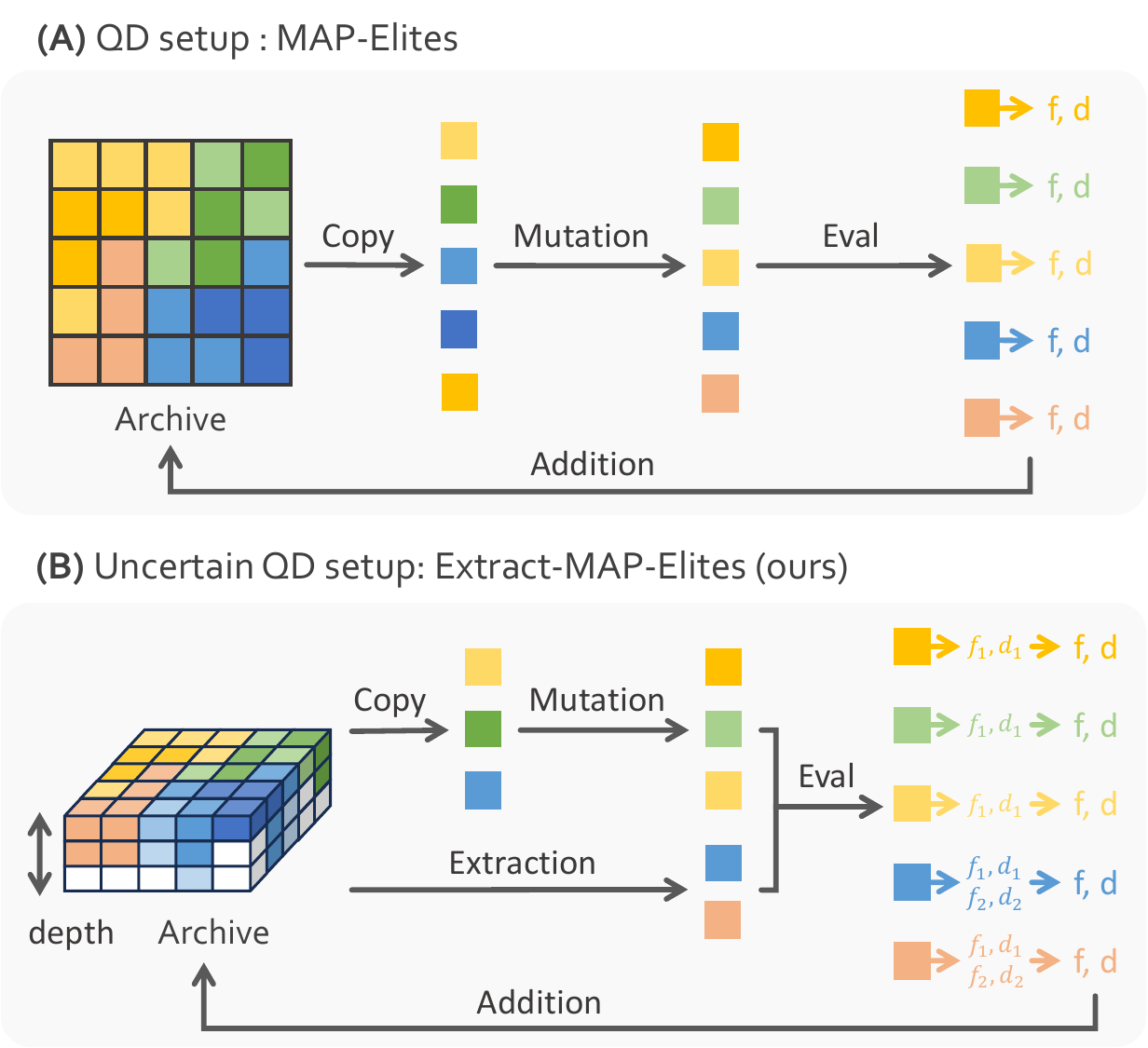}
  \caption{
    Comparison of (A) the existing ME algorithm with (B) our proposed \name{} algorithm, which is designed to address the uncertainty present in UQD tasks. 
    \name{} introduces an extraction mechanism that re-evaluates solutions within the archive, an estimation mechanism and a depth to the archive. 
  }
  \Description{Illustration of the difference between our proposed \name{} and MAP-Elites}
  \label{fig:algo}
\end{figure}

\name{} proposes to dedicate a fixed proportion (here $25\%$) of the evaluations per generation to re-evaluating elites and leave the rest to new offspring. 
For example, if ME evaluates $128$ offspring per generation, \name{} would only generate and evaluate $96$ new offspring, and use the remaining $32$ evaluations to re-evaluate elites already in the archive. 
When selected to be re-evaluated, elites are "extracted" from the archive, meaning that they are removed from the archive. This is unlike parents chosen to create new offspring that are typically "copied" from the archive. 
As detailed in Section~\ref{sec:adaptive_sampling}, re-evaluated elites tend to "drift". i.e. prove to belong to another cell than the one they were originally in. The extraction in \name{} accounts for this by concatenating extracted elites and new offspring into an array of solutions that can be evaluated and added to the archive together following the same rules, as illustrated in Figure~\ref{fig:algo}. 
In \name{}, elites are extracted randomly from the archive to be re-evaluated, following a probability law detailed in Section~\ref{sec:depth}.

\subsubsection{Performance Estimation} 
In \name{}, each solution has a buffer of fitness and descriptor feature evaluations, used to store all its evaluations, whether they come from its first evaluation or from re-evaluations. 
The evaluation process concatenates new evaluations to this buffer as shown in Figure~\ref{fig:algo}, accumulating information that is used to infer better and better estimates of the performance of elites as they get extracted and re-evaluated. 

\subsubsection{Archive Depth} \label{sec:depth}
\name{} maintains a depth in each cell that contains the $d$ more promising elites as illustrated in Figure~\ref{fig:algo} (we use $d=8$). Only the top layer of this depth is returned as the final archive, the rest of the depth is only used during optimisation, as a memory buffer of promising solutions. Elites in the depth might move to the top when they get re-evaluated or replace an extracted elite.
In \name{}, elites are extracted from the depth with a probability exponentially-proportional to their rank in their cell: the higher in the depth of the cell a solution is, the more likely it would get re-evaluated. Using exponential (instead of linear, for example) enables using a significantly larger depth $d$ while maintaining a higher probability to extract the best-performing solutions (i.e. maintains extraction pressure). 
At each generation of \name{}, the extraction phase leaves empty slots in each cell, which are filled by shifting upward the solutions stored lower in the depth.

\subsubsection{Relation to AS and Adapt-ME} \name{} is an adaptive-sampling approach, like AS and Adapt-ME (see Section~\ref{sec:adaptive_sampling}), this section details how they compare.
Unlike AS, \name{} adapts to the number of evaluations available per generation instead of re-evaluating the whole archive content. Lifting this constraint is crucial as it makes \name{} independent of the total archive size and allows it to generalise to a wide range of domains with large archives or limited evaluations per generation. 
Additionally, \name{} distributes its evaluation more wisely than AS, as it has a higher chance to re-evaluate solutions higher in the depth, while AS re-evaluates all solutions in the depth equally. 
Unlike Adapt-ME, \name{} is parallelisable and has a fixed per-generation evaluation budget, drastically reducing its run time. 
Additionally, \name{} re-evaluates any elites in the archive and not only the ones that get challenged by offspring, thus improving performance estimation across the descriptor space. 

\subsection{\Longframework{}} \label{sec:framework}

The \framework{} proposes a general modular view of all existing UQD approaches introduced in Section~\ref{sec:uqd}, including our proposed \name{}. 
It builds on top of the QD-Framework from Section~\ref{sec:qd_framework}, as illustrated in Figure~\ref{fig:framework}. 
In this section, we first introduced its modules, before detailing how it encompasses previous approaches and how it enables building new UQD approaches. 

\subsubsection{Modules} 
\begin{figure}[t!]
  \centering
  \includegraphics[width=\linewidth]{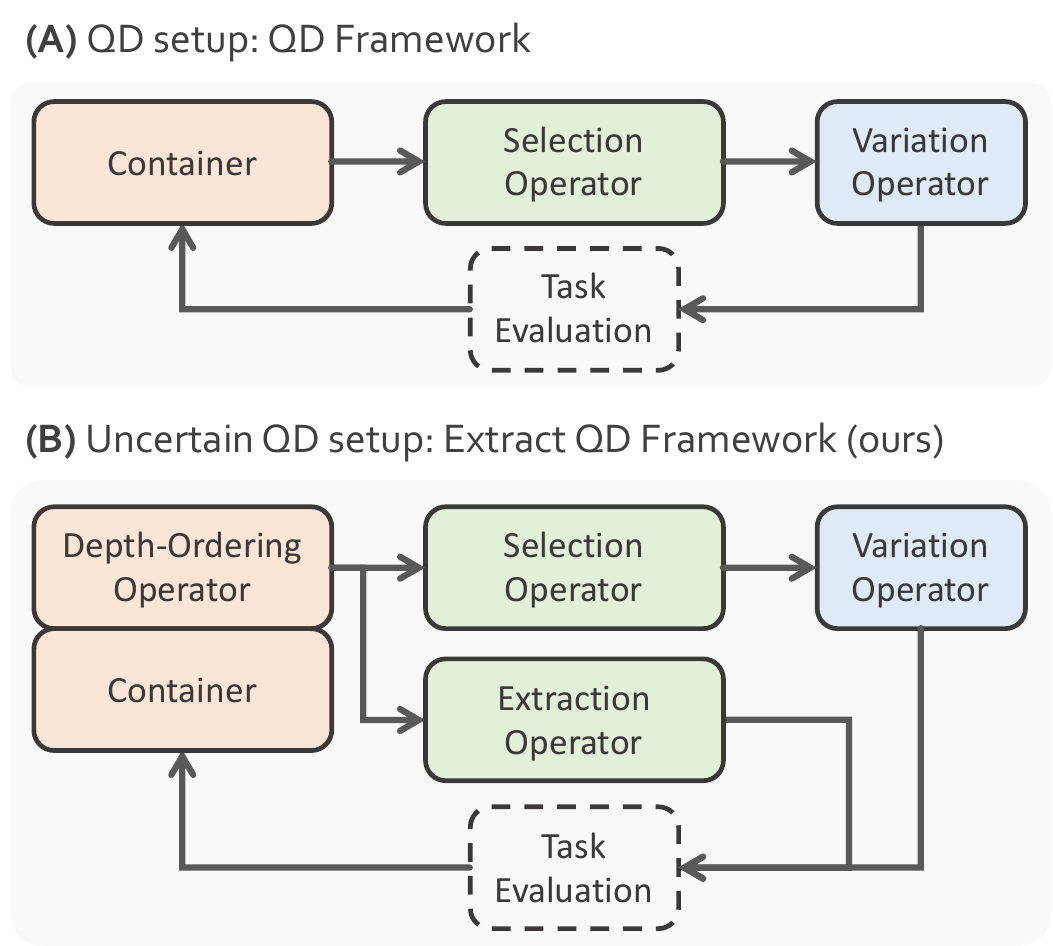}
  \caption{
    Illustration of (A) the existing QD Framework for the QD setup, and (B) our proposed \framework{} that extends it for uncertain tasks. The \framework{} encompasses existing UQD approaches and allows practitioners to easily build new tailored approaches for UQD tasks.
  }
  \Description{Illustration of the difference between our proposed \framework{} and the QD-Framework}
  \label{fig:framework}
\end{figure}

The \framework{} comprises the following modules, also illustrated in Figure~\ref{fig:framework}:

\begin{itemize}
    \item \textbf{Selection Operator:} as in the original QD Framework, this operator selects parents to mutate new offspring solutions. Possible Selection Operators include uniform selection, fitness-proportional selection, biased wheel, or random tournament~\cite{framework}. 

    \item \textbf{Variation Operator:} this operator was present in the original QD Framework, though not explicitly identified as a module. Recent developments in QD algorithms have focused on modifying this operator~\cite{tjanaka2022approximating, pierrot2022diversity, nilsson2021policy, faldor2023synergizing, flageat2024enhancing}, making it crucial in algorithm design. Thus, we include it as a distinct module in both the QD Framework and the \framework{} in Figure~\ref{fig:framework}. It applies mutation to evolve parent solutions into offspring. It includes polynomial or Gaussian random mutations as well as more complex mechanisms such as gradient-based mutations. 

    \item \textbf{Extraction Operator:} at each generation, part of the evaluations are spent re-evaluating elites that are extracted from the container (as in \name{}, see Section~\ref{sec:extraction}). The Extraction Operator chooses which and how many elites are re-evaluated. 
    It includes cases where elites are extracted randomly, as in \name{}, as well as cases where a fixed deterministic subset of elites is extracted, such as in AS where the entire population is extracted (Section~\ref{sec:adaptive_sampling}). 
    The concatenation of the solutions output by the Variation and Extraction Operators forms the overall population to evaluate, as shown in Figure~\ref{fig:framework}. 

    \item \textbf{Container with Depth:} the \framework{} keeps the concept of container from the original QD Framework but proposes to augment it with a depth $d$. The container organizes solutions into an ordered collection that covers the feature descriptor space and constructs the final collection. Adding depth $d$ involves maintaining a memory buffer of $d$ promising elites for each neighbourhood of the feature descriptor space. Thus, a depth of $d=1$ corresponds to the container of the standard QD Framework. 
    This paper focuses on ME-based algorithms, where the feature descriptor neighbourhoods are the cells of the grid, so the depth is directly added to the cells as in \name{}. 
    However, this can be extended to other types of containers. 
    In NSLC-based approaches, the container structure autonomously emerges from the encountered solutions, defining neighbourhoods~\cite{lehman2011evolving}. These neighbourhoods are updated when encountering better solutions, so they can maintain $d$ solutions instead of a single one. 
    For multi-objective QD that maintain pareto-fronts~\cite{pierrot2022multi, janmohamed2023improving}, the depth can maintain multiple Pareto fronts via Pareto ranking, as done in similar works in EC~\cite{siegmund2013comparative, siegmund2015hybrid}. 
    
    \item \textbf{Depth-Ordering Operator:} this operator ranks the elites within the depth of the container. Thus, it defines which elites belong to the "top" layer and are returned as part of the final collection. 
    This operator enables the incorporation of more complex mechanisms than simply fitness-based ranking. For example, previous UQD works~\cite{flageat2024exploring} have used a combination of fitness and reproducibility to order solutions within the depth, not necessarily returning the highest-fitness one.

    \item \textbf{Samples:} the number of samples spent on the first evaluation of solutions is an important parameter in uncertain tasks, so we include it as a framework's parameter. 

\end{itemize}

\input{Tables/appraoches_table}

\subsubsection{UQD Approaches in the \framework{}}
A key strength of our \framework{} is its ability to encompass existing UQD approaches, as shown in Table~\ref{tab:approaches}. This section details the most common approaches, we provide the others in Appendix~\ref{app:framework}.

Although \textbf{ME}~\cite{map_elites} is not a UQD method, we begin by showing its compatibility with our framework, as it forms the foundation of all other approaches.
ME utilizes a grid-based container and applies uniform selection over the grid as its selection operator, consistent with the original QD Framework~\cite{framework}. ME also has a depth of $1$, does not use extraction (i.e., number of extracts of $0$ with any extraction operator), and retains only the highest-fitness solution in each cell (i.e., depth-ordering operator based solely on fitness). 
\textbf{ME-Sampling}~\cite{adaptive}, the most widely spread UQD approach, is identical to ME but uses $32$ samples for each solution instead of $1$. Other fixed-sampling approaches such as \textbf{ME-Low-Spread}~\cite{mace2023quality} or \textbf{ME-Weighted}~\cite{flageat2024exploring} simply swap the depth-ordering operator to use a mixture of fitness and reproducibility to choose the best solutions. 
\textbf{MOME-X}~\cite{flageat2024exploring} swaps the grid container for a MOME container~\cite{pierrot2022multi} that maintains a Pareto front of solutions within each cell. 
\textbf{Deep-Grid}~\cite{flageat2020fast} uses a depth of $32$ solutions in each cell and fitness-proportional selection operator within this depth.
It also uses a depth-ordering operator based on seniority and fitness, in which only the latest solutions are kept within the depth, and ordered based on fitness.
\textbf{AS}~\cite{flageat2023uncertain} also uses depth and is characterised by a periodic re-evaluation of all elites in the archive that corresponds to an extraction operator that extracts the full content of the archive.

\subsubsection{Building New UQD Approaches using the \framework{}}
The \framework{} also aims to identify the modular components that can be swapped to design new UQD approaches. These modules can be customised using various types of knowledge, such as task-specific insights.
Consider a task where the distributions of fitness and descriptors are known, this knowledge can be utilised to design improved depth-ordering or extraction operators, for example, based on confidence as done in EC~\cite{lecarpentier2022lucie}.
Alternatively, if the mapping from genotype to descriptor and fitness is known to be smooth, neighbouring solutions in the genotype space can be used to approximate descriptors and fitness values, reducing the need for evaluations, as done in similar work in EC~\cite{ea_archive}.
Similarly, our \framework{} enables incorporating uncertainty-handling mechanisms with existing QD approaches. 
In our experiments, we propose Extract-PGA (EPGA), a variant of \name{} that swaps the Variation Operator module to use the variation from the widely-spread PGA-ME algorithm~\cite{nilsson2021policy}, adapting it for uncertain tasks.

%% file: Tables/appraoches_table.tex
\begin{table*}[t!]
\small
  \caption{
    Overview of how UQD approaches fit within our \framework{}.
    $b$ denotes the number of offspring per generation, $C$ the number of cells in the grid, $d$ the depth and $N$ the number of first evaluation samples. Reprod stands for reproducibility.
  }
  \begin{tabular}{ c | c | c c c | c c | c c }

    & \textsc{\makecell{Samples \\ $N$}}
    & \textsc{\makecell{Container}} 
    & \textsc{\makecell{Depth \\ $d$}} 
    & \textsc{\makecell{Depth-Ordering \\ Operator}} 
    & \textsc{\makecell{Selection \\ Operator}} 
    & \textsc{\makecell{Variation \\ Operator}} 
    & \textsc{\makecell{Extraction \\ Operator}} 
    & \textsc{\makecell{Number \\ Extracts}}
    \\
    
    \midrule

    \textsc{\makecell{ME \cite{map_elites}}}
    & $1$
    & \makecell{Grid of $C$ cells}
    & $1$
    & \makecell{Fitness rank}
    & Uniform
    & Mutation
    & -
    & $0$
    \\

    \midrule
    
    \textsc{\makecell{ME-Sampling \cite{adaptive}}}
    & $32$
    & \makecell{Grid of $C$ cells}
    & $1$
    & \makecell{Fitness rank}
    & Uniform
    & Mutation
    & -
    & $0$
    \\
    
    \textsc{\makecell{AS \cite{flageat2023uncertain}}}
    & $2$
    & \makecell{Grid of $C$ cells}
    & $2$
    & \makecell{Fitness rank}
    & \makecell{Uniform, \\ top of depth}
    & Mutation
    & \makecell{Grid content}
    & $Cd$
    \\
    
    \textsc{\makecell{Deep-Grid \cite{flageat2020fast}}}
    & $1$
    & \makecell{Grid of $C$ cells}
    & $32$
    & \makecell{Latest to enter \\ fitness ranked}
    & \makecell{Uniform, \\ depth fit-prop}
    & Mutation
    & -
    & $0$
    \\
    
    \textsc{\makecell{Adapt-ME \cite{adaptive}}}
    & $1$
    & \makecell{Grid of $C$ cells \\ + buffer of $b$}
    & $8$
    & \makecell{Fitness rank}
    & \makecell{Uniform, \\ top of depth}
    & Mutation
    & \makecell{Subset of \\ buffer and grid}
    & \makecell{$\leq b + bd$}
    \\

    \midrule

    \textsc{\makecell{ME-Reprod \cite{grillotti2023don}}}
    & $32$
    & \makecell{Grid of $C$ cells}
    & $1$
    & \makecell{Reprod rank}
    & Uniform
    & Mutation
    & -
    & $0$
    \\

    \textsc{\makecell{ME-Weighted \cite{flageat2024exploring}}}
    & $32$
    & \makecell{Grid of $C$ cells}
    & $1$
    & \makecell{Sum of fitness \\ and reprod rank }
    & Uniform
    & Mutation
    & -
    & $0$
    \\
    
    \textsc{\makecell{AS-Weighted  \cite{flageat2024exploring}}}
    & $2$
    & \makecell{Grid of $C$ cells}
    & $2$
    & \makecell{Sum of fitness \\ and reprod rank}
    & \makecell{Uniform, \\ top of depth}
    & Mutation
    & \makecell{Grid content}
    & $Cd$
    \\

     \textsc{\makecell{ME-Low-Spread \\ \cite{mace2023quality}}}
    & $32$
    & \makecell{Grid of $C$ cells}
    & $1$
    & \makecell{Improvement in \\ fitness and reprod}
    & Uniform
    & Mutation
    & -
    & $0$
    \\

    \midrule

    \textsc{\makecell{ME-Delta \cite{flageat2024exploring}}}
    & $32$
    & \makecell{Grid of $C$ cells}
    & $1$
    & \makecell{Dominance in \\ fitness and reprod}
    & Uniform
    & Mutation
    & -
    & $0$
    \\
    
    \textsc{\makecell{AS-Delta \cite{flageat2024exploring}}}
    & $2$
    & \makecell{Grid of $C$ cells}
    & $2$
    & \makecell{Dominance in \\ fitness and reprod}
    & \makecell{Uniform, \\ top of depth}
    & Mutation
    & \makecell{Grid content}
    & $Cd$
    \\

    \textsc{\makecell{MOME-X \cite{flageat2024exploring}}}
    & $32$
    & \makecell{Grid of $C$ \\ Pareto fronts}
    & $1$
    & \makecell{Pareto front of \\ fitness and reprod}
    & \makecell{Crowding \\ distance based}
    & Mutation
    & -
    & $0$
    \\

    \midrule

    \textsc{\makecell{ARIA \cite{grillotti2023don}}}
    & $1024$
    & \makecell{Grid of $C$ cells}
    & $1$
    & \makecell{Fitness rank}
    & \makecell{Edge of \\ filled cells}
    & \makecell{Constrained \\ fitness on reprod}
    & -
    & $0$
    \\

    \midrule

    \textsc{\makecell{\name{} (ours)}}  
    & $2$
    & \makecell{Grid of $C$ cells}
    & $8$
    & \makecell{Fitness rank}
    & \makecell{Uniform \\ top of depth}
    & Mutation
    & \makecell{$pb$ sampled \\ elites}
    & \makecell{$min(pb, C)$}
    \\

  \end{tabular}
  \label{tab:approaches}
\end{table*}

%% file: Parts/4_Experiments.tex
\section{Experimental Study}

We propose two experimental studies. First, we demonstrate the capability of our proposed \name{} method to generate collections of diverse solutions in uncertain domains. Second, we illustrate the benefits of our proposed \framework{} by showing how it improves the performance of the widely used PGA-ME algorithm~\cite{nilsson2021policy} in uncertain domains at no additional evaluation cost.

\subsection{Experimental study of \Longname{}} \label{sec:results_algo}


\subsubsection{Tasks}
We consider the following tasks (Table~\ref{tab:tasks}), used in previous UQD work~\cite{flageat2023uncertain, flageat2024exploring, grillotti2023don} (more details in Appendix~\ref{app:setup}):

\input{Tables/tasks_table}

\begin{figure*}[t!]
  \centering
  \includegraphics[width=\linewidth]{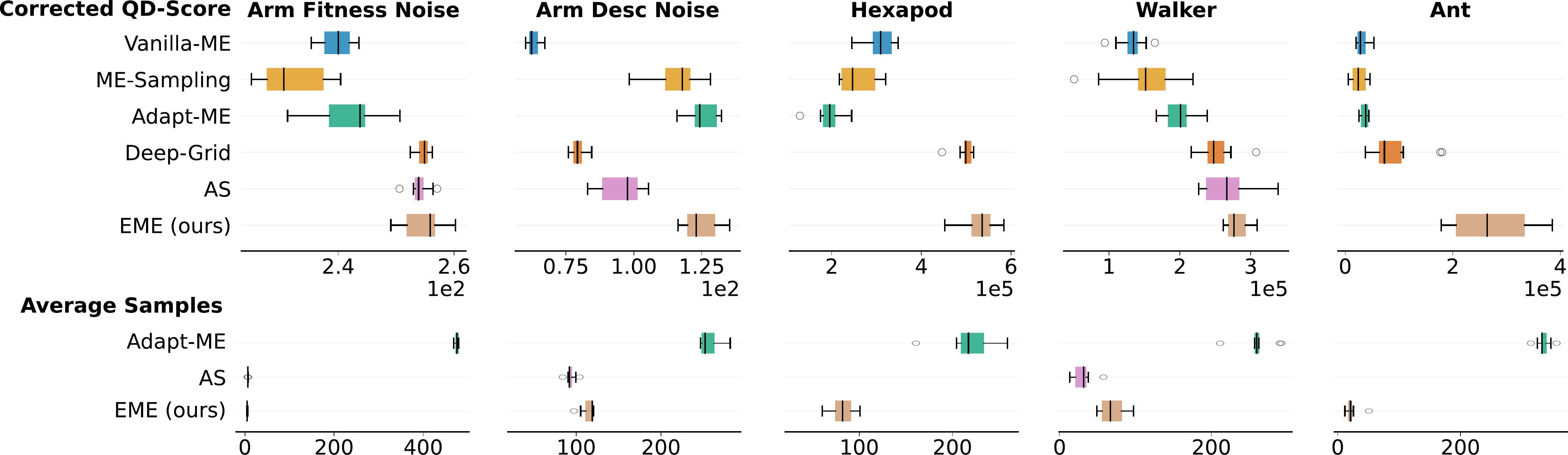}
  \caption{
    Comparison of our \name{} approach with similar existing UQD approaches. The Corrected QD-Score reflects performance on the task, while the Average Samples offer insights into how the methods manage the available sampling budget. The vertical line represents the median across $10$ replications; the box shows the quartiles, the whiskers indicate $1.5$ times the interquartile range, and the dots represent outliers. Blank lines indicate undefined approach-task pairs.
  }
  \label{fig:main_results}
  \Description{Main comparison results}
\end{figure*}

\subsubsection{Baselines} We use the following baselines from Section~\ref{sec:uqd}: (1) Vanilla-ME, (2) ME-Sampling, (3) Adapt-ME, (4) AS, and (5) Deep-Grid. We use the parameters from previous works~\cite{flageat2023uncertain, flageat2024exploring}. We replicate each algorithm-task pair across $10$ seeds. 
All our implementations use the QDax library~\cite{chalumeau2024qdax, lim2022accelerated}.
To facilitate later works, we open-source our code at \textit{URL-to-be-released-upon-acceptance}.

\subsubsection{Metrics}
A key challenge in UQD is that the archive returned by an algorithm is not trustworthy. For instance, when ME uses only a single evaluation per solution, the resulting archive is likely to be filled with misplaced "lucky" solutions. Thus, the archive returned by an algorithm is referred to as an \textbf{illusory archive}, and it is usually adjusted into a \textbf{corrected archive}~\cite{adaptive}, used to compute metrics. To construct the corrected archive, we add all the solutions from the illusory archive in an empty archive (with no depth) using their "ground-truth" fitness and descriptor feature. 
When the ground-truth is not directly accessible, we approximate it as the median of $512$ re-evaluations (shown to be a reasonable number in previous work~\cite{flageat2023uncertain}). 
Similar to previous work, we then report the \textbf{Corrected QD-Score}: the sum of all fitness in the corrected archive. It quantifies the overall performance and diversity of the final collection. 
Additionally, to better understand the dynamic of \name{}, we report the \textbf{Average Samples} spent on solutions in the illusory archive (only in the top layer if there is a depth). 
We report $p$-values using the Wilcoxon test with Holm-Bonferroni correction. 

\begin{figure}[t!]
  \centering
  \includegraphics[width=0.95\linewidth]{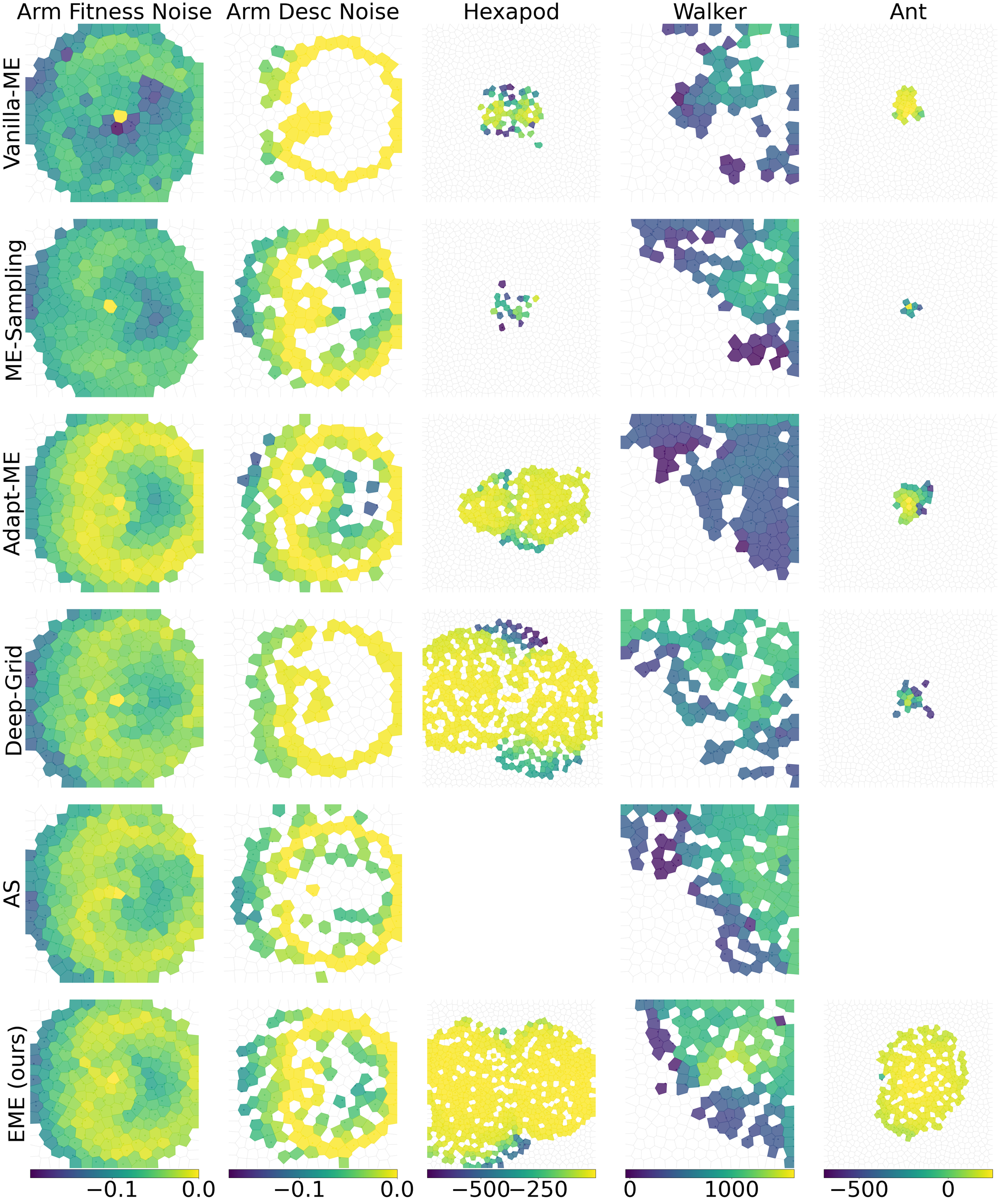}
  \caption{
    Final Corrected Archive of a random replication of each approach across all tasks. Blank panels indicate undefined approach-task pairs. 
    Each cell corresponds to a solution, with brighter colours indicating higher fitness. The axes represent the descriptor dimensions detailed in Table~\ref{tab:tasks}.
  }
  \Description{Final archive comparison plots}
  \label{fig:archives}
  \vspace{-1mm}
\end{figure}

\subsubsection{Sampling-Size Comparison} \label{sec:sampling_size}
UQD algorithms are typically compared using a fixed maximum number of evaluations per generation, referred to as sampling-size~\cite{flageat2023uncertain}.
For instance, with a sampling-size of $1024$, ME produces $1024$ offspring per generation, evaluated $1$ time each, while ME-Sampling generates only $32$ offspring per generation, each evaluated $32$ times, giving also a total of $1024$ evaluations. All our comparisons use a sampling size of $1024$.

\subsubsection{Results}

We report the results of our experiments in Figure~\ref{fig:main_results} and the corresponding final corrected archives in Figure~\ref{fig:archives}. 
\textbf{Vanilla-ME} performs among the worst across the board which is consistent with the fact that it does not account for uncertainty. 
\textbf{ME-Sampling} performs well on tasks with descriptor noise but its tendency to hinder exploration is visible on more complex control tasks such as Ant.  
\textbf{Deep-Grid} is a strong competitor in tasks with limited descriptor noise but, by construction, it struggles when the noise on the descriptor becomes too large compared to the size of the cells (Arm Desc Noise and Ant tasks). 
\textbf{AS} is not defined in two of the tasks (Hexapod and Ant) because re-evaluating everything in the archive requires too many evaluations for the available budget, showcasing the limitations of this approach. However, when defined, it is one of the best-performing approaches. 
\textbf{Adapt-ME} performs well on the low-dimensional Arm task but struggles with higher-dimensional tasks. The Average Samples indicate that a significant portion of the sampling budget is spent on re-evaluating solutions, which severely hampers its exploration capability. 
Finally, our proposed \textbf{\name{}} performs well across the board, matching the performance of the best approach for each task and even outperforming all other approaches on Ant ($p<10^{-3}$). 
Average Samples indicate that \name{} uses significantly more samples than AS on the elites in the final archive for Arm Desc Noise and Walker ($p<10^{-3}$). This highlights that the extraction mechanism achieves a better allocation of samples compared to re-evaluating the entire archive.
More broadly, our results show the wide applicability of \name{} and its strength as a "first guess" method for new UQD tasks.

\subsection{Benefits of the \Longframework{}: Example of PGA-ME} \label{sec:results_qdrl}

Quality-Diversity Reinforcement Learning (QD-RL) algorithms~\cite{tjanaka2022approximating, pierrot2022diversity, nilsson2021policy, faldor2023synergizing, tjanaka2023training} are QD algorithms specifically designed for Reinforcement Learning (RL) tasks. The properties available in these RL tasks enable the development of efficient targeted QD algorithms. 
Surprisingly, while most RL environments are uncertain (e.g. random initialisation or stochastic transitions), the aspects of UQD are rarely considered in the QD-RL literature.
Here, we propose to combine a common QD-RL approach: Policy-Gradient-Assisted ME (PGA)~\cite{nilsson2021policy} within our \framework{}. 
By doing so, we aim to illustrate how accounting for uncertainty and integrating UQD insights can improve the performance of common QD algorithms.

\subsubsection{Algorithm} 
We propose \Longnamepga{} (\namepga{}), a variant of PGA using our proposed \framework{}. 
The original PGA algorithm builds on ME, replacing its variation operator with RL variations based on the Twin Delayed Deep Deterministic (TD3)~\cite{fujimoto2018addressing} algorithm. Similarly, \namepga{} corresponds to our \name{} algorithm swapping the variation operator to use the one from PGA.

\subsubsection{Tasks}
We consider three tasks of the QD-Gym suite proposed in the original PGA paper~\cite{nilsson2021policy}, detailed in Table~\ref{tab:tasks_qdrl}.

\input{Tables/tasks_qdrl_table}

\subsubsection{Baselines and Metrics}
We compare PGA, ME and \namepga{} for $128$ evaluations per generation. While ME and PGA generate $128$ new solutions, \namepga{} only generates $96$ new solutions and uses the remaining $32$ evaluations to re-evaluate elites from the archive. 
We report the final Corrected QD-Score for each algorithm, as in the previous experimental section in Section~\ref{sec:results_algo}. 

\subsubsection{Results}

\begin{figure}[t!]
  \centering
  \includegraphics[width=\linewidth]{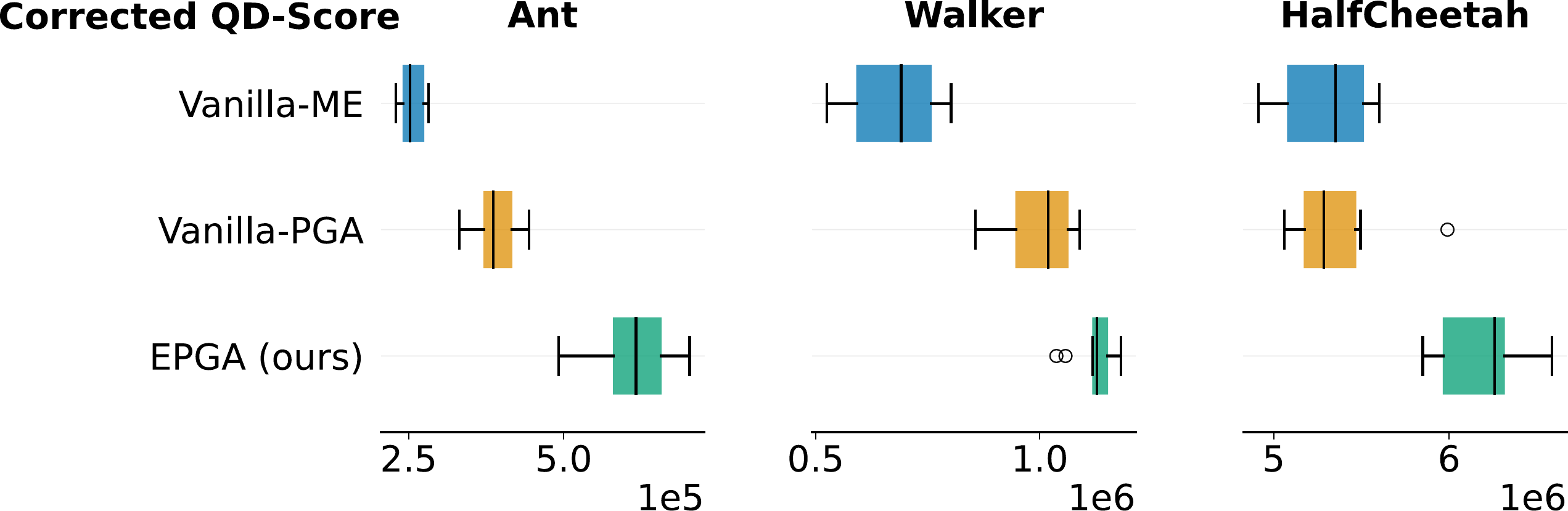}
  \caption{
    Comparison of our proposed \namepga{}{} approach with Vanilla-ME and PGA in terms of final Corrected QD-Score. The vertical line represents the median across $10$ replications; the box shows the quartiles, the whiskers indicate $1.5$ times the interquartile range, and the dots represent outliers. 
  }
  \label{fig:qdrl_results}
  \Description{QDRL comparison results}
  \vspace{-4mm}
\end{figure}

We report the results of our experiments in Figure~\ref{fig:qdrl_results}. 
Our proposed \namepga{} approach, built with our \framework{} outperforms the PGA algorithm across all tasks ($p<5.10^{-3}$). 
Crucially, this improvement is done at no additional evaluation cost: all approaches are evaluated for the same total number of evaluations and with the same number of evaluations per generation (i.e. sampling size). 
In \namepga{}, re-evaluating elites enables accounting for uncertainty and building better estimation of the fitness and feature descriptors of solutions, allowing the method to generate higher-quality archives. These results highlight the importance of accounting for uncertainty in QD tasks and demonstrate the value of our \framework{} in developing better QD algorithms.

%% file: Tables/tasks_table.tex
\begin{table}[ht]
\centering
\small 
  \caption{Task suite considered in this work.}
  \begin{tabular}{ c | c c c c }

    & \makecell{\textsc{Arm} \\ \includegraphics[width = 0.06\textwidth]{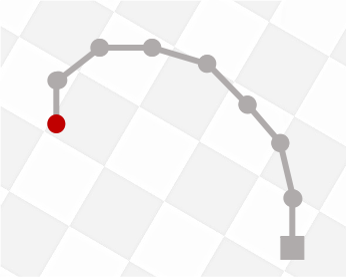}}
    & \makecell{\textsc{Hexapod} \\ \includegraphics[width = 0.06\textwidth]{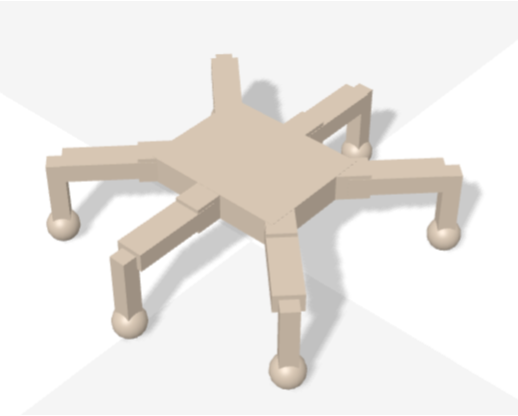}}
    & \makecell{\textsc{Walker} \\ \includegraphics[width = 0.06\textwidth]{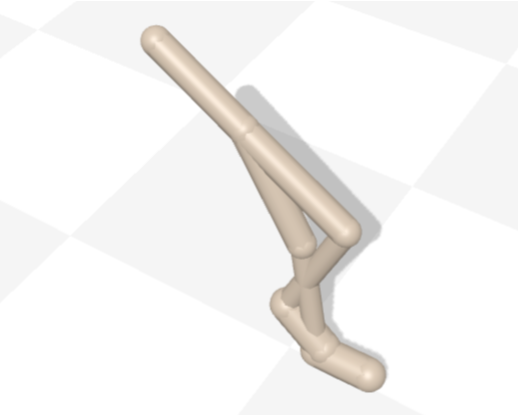}}
    & \makecell{\textsc{Ant} \\ \includegraphics[width = 0.06\textwidth]{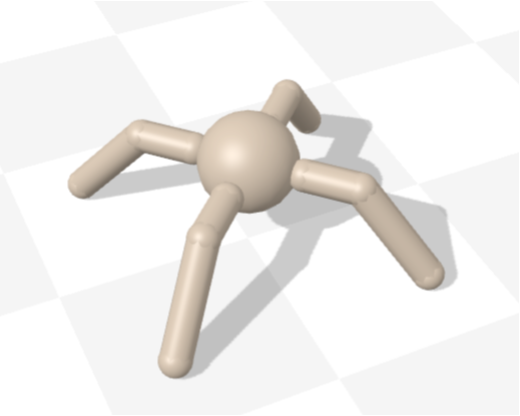}}
    \\

    \midrule

    \textsc{Control} 
    & \makecell{Joints \\ position}
    & \makecell{Periodic \\ functions}
    & \makecell{Neural \\ network}
    & \makecell{Neural \\ network}
    \\

    \textsc{Dims} 
    & $\mathit{8}$
    & $\mathit{24}$
    & $\mathit{270}$
    & $\mathit{312}$
    \\

    \midrule

    \textsc{Fitness} 
    & \makecell{- Joints \\ position \\ variance}
    & \makecell{Direction \\ error}
    & \makecell{+ Speed \\ - energy \\ + survival}
    & \makecell{- Energy \\ + survival}
    \\

    \midrule
    
    \textsc{\makecell{Descriptor}} 
    & \makecell{End-effector \\ position}
    & \makecell{Final \\ position}
    & \makecell{Feet \\ contact}
    & \makecell{Final \\ position}
    \\


    \textsc{Dims} 
    & $\mathit{2}$
    & $\mathit{2}$
    & $\mathit{2}$
    & $\mathit{2}$
    \\

    \textsc{Nb Cells} 
    & $\mathit{256}$
    & $\mathit{1024}$
    & $\mathit{256}$
    & $\mathit{1024}$
    \\

    \midrule
    
    \textsc{\makecell{Uncertainty}} 
    & \makecell{Fitness OR \\ Descriptor}
    & \makecell{Fitness and \\ Descriptor}
    & \multicolumn{2}{c}{\makecell{Initial joints' position \\ and velocity}}
    \\
  
  \end{tabular}
  \label{tab:tasks}
\end{table}

%% file: Tables/tasks_qdrl_table.tex
\begin{table}[ht]
\centering
\small
  \caption{Task suite considered for the QDRL tasks.}
  \begin{tabular}{ c | c c c }

    & \makecell{\textsc{Walker} \\ \includegraphics[width = 0.06\textwidth]{Figures/walker.png}}
    & \makecell{\textsc{Cheetah} \\ \includegraphics[width = 0.06\textwidth]{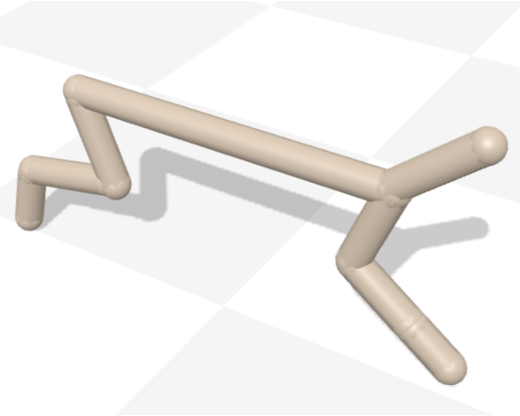}}
    & \makecell{\textsc{Ant} \\ \includegraphics[width = 0.06\textwidth]{Figures/ant.png}}
    \\

    \midrule

    \textsc{Control} 
    & \multicolumn{3}{c}{Neural network}
    \\

    \textsc{Dims} 
    & $\mathit{5702}$
    & $\mathit{5766}$
    & $\mathit{6600}$
    \\

    \midrule

    \textsc{Fitness} 
    & \multicolumn{3}{c}{+ speed - energy + survival}
    \\

    \midrule
    
    \textsc{Descriptor} 
    & \multicolumn{3}{c}{Feet contact with the ground}
    \\


    \textsc{Dims} 
    & $\mathit{2}$
    & $\mathit{2}$
    & $\mathit{4}$
    \\

    \textsc{Nb Cells} 
    & $\mathit{1024}$
    & $\mathit{1024}$
    & $\mathit{1024}$
    \\

    \midrule
    
    \textsc{Uncertainty} 
    & \multicolumn{3}{c}{Initial joints' position and velocity}
    \\
  
  \end{tabular}
  \label{tab:tasks_qdrl}
\end{table}

%% file: Parts/5_Conclusion.tex
\section{Discussion and Conclusion}

This work focuses on QD algorithms applied to uncertain domains and introduces two main contributions: the \framework{}, a generic framework that encompasses existing UQD approaches and facilitates the design of new ones, and \name{}, a new UQD algorithm that directly instantiates this framework.
These contributions are intended as tools for addressing new UQD tasks: \name{} provides a reliable "first guess" method, likely to perform well, while \framework{} serves as a toolbox for developing tailored UQD approaches for specific tasks or accounting for uncertainty in existing QD methods.
We present an experimental study demonstrating the strong performance of \name{} on widely used UQD tasks. Additionally, we demonstrate the capability of our \framework{} to easily create new and high-performing UQD algorithms, showing that accounting for uncertainty and integrating UQD insights can significantly enhance the performance of standard algorithms.

A limitation of this work is that it focuses on the Performance Estimation problem, without addressing the two other known UQD problems, namely Reproducibility Maximisation and Fitness-Reproducibility Trade-off. While this choice was made for clarity, we demonstrate in Section~\ref{sec:framework} and Table~\ref{tab:approaches} that our \framework{} encompasses existing approaches that account for these two problems. This also highlights the potential of the framework for deriving new approaches that address these two problems, although a detailed study of this is left for future work.

This work consolidates existing knowledge from previous UQD research while advancing our understanding of how existing methods can be adapted to novel UQD tasks. Our contributions aim to lower the entry barrier to UQD methods and insights, making an important step to extend the applicability of QD algorithms.

%% file: Parts/Acknowledgement.tex
We would like to thank Paul Templier and Hannah Janmohamed from the Adaptive and Intelligent Robotics Lab for their valuable inputs, and the members of the QDGrasp team from ISIR for their feedback. This work was supported by the German Ministry of Education and Research (BMBF) (01IS21080), the French Agence Nationale de la Recherche (ANR) (ANR-21-FAI1-0004) (Learn2Grasp), the European Commission's Horizon Europe Framework Programme under grant No 101070381 and from the European Union's Horizon Europe Framework Programme under grant agreement No 101070596.

%% file: Parts/Appendix.tex

\section{Additional Details on UQD Approaches in the \framework{}} \label{app:framework}

This Section provides additional details on how each UQD approach fits within our proposed \framework{}. 

\paragraph{ME-Sampling}
As detailed in Section~\ref{sec:framework}, ME-Sampling (see Section~\ref{sec:fixed_sampling}) is exactly similar to ME except that it uses $32$ samples per solution during their first evaluation. Thus, ME-Sampling does not use any extraction or mechanisms that can be assimilated. In other words, in ME-Sampling, once elites have been evaluated and added to the archive, they cannot be re-evaluated or re-questioned.

\paragraph{Deep-Grid}
As mentioned in Section~\ref{sec:framework}, Deep-Grid~\cite{flageat2020fast} (see Section~\ref{sec:deep_grid}) differs from ME by using a depth of $32$ solutions in each cell and fitness-proportional selection within this depth.
Additionally, in Deep-Grid, new solutions can replace any solutions already in the cell, no matter the results of their evaluations, and only the highest-fitness ones are returned as the final collection.
This corresponds to using Depth-Ordering based on seniority and fitness, in which only the latest solutions are kept within the depth, and these solutions are ordered based on fitness.
Deep-Grid does not use any extraction operator, meaning that elites never get explicitly re-evaluated. Instead, Deep-Grid relies on the assumption that, as elites are selected as parents, they generate offspring close to them, leading to indirectly re-evaluating their area of the search space.

\paragraph{Archive-Sampling (AS)}
As detailed in Section~\ref{sec:framework}, AS~\cite{flageat2023uncertain} (see Section~\ref{sec:adaptive_sampling}) uses the same container and selection operator as ME. Its container is a grid with depth $d$, ordered based on fitness, i.e. only the $d$ best-performing solutions are kept within each cell. AS also uses an extraction operator that extracts the full content of the archive at each generation, depth included, thus emptying it fully. All solutions in the archive are re-evaluated with the new offspring before being added back into the archive. Thus, the number of extracts in AS is equal to the total size of the container: $Cd$ where $C$ is the number of cells in the grid and $d$ is the depth.

\paragraph{Adapt-ME}
Adapt-ME~\cite{adaptive} (see Section~\ref{sec:adaptive_sampling}) uses the same container and selection operator as ME. However, it re-evaluates part of the elites, who are chosen as the ones a new offspring has challenged. This behaviour can easily be modelled as an extraction. Additionally, Adapt-ME requires offspring to be evaluated the same number of times as the elites in their respective cells. 
We propose to express this using a container composed of a grid and a buffer of the offspring that still needs re-evaluation. At each generation, solutions are extracted from both the grid and the buffer using the Extraction Operator, unifying these processes into a single extraction mechanism.
This new perspective on Adapt-ME enables better parallelisation of evaluations compared to the original algorithm, where evaluations were performed sequentially. However, it does not address the issue that the number of extracted solutions varies between generations. Thus, we can only provide an upper bound for this quantity: $b + bd$, where $b$ represents the number of offspring per generation and $d$ is the depth. If all offspring require an additional evaluation before addition, $b$ additional offspring from the buffer would need to be re-evaluated, and if all offspring trigger elite re-evaluation, $bd$ elites would be extracted from the grid.

\paragraph{ME-Reprod}
ME-Reprod~\cite{grillotti2023don} is a variant of ME-Sampling that uses the $32$ samples to compute reproducibility and only optimises for this. Thus, all its components are the same as ME-Sampling except that the Depth-ordering operator is based on reproducibility only instead of fitness. ME-Reprod is commonly used as a baseline to estimate an upper bound on reproducibility~\cite{grillotti2023don, flageat2024exploring}. However, due to the Performance-Reproducibility Trade-off (the third problem in UQD, see Section~\ref{sec:problems}), the reproducibility values achieved by ME-Reprod are often unattainable for high-performing solutions. For instance, in robotics, controllers that remain stationary can reach reproducibility values that are out of reach for moving controllers.

\paragraph{ME-Weighted}
ME-Weighted~\cite{flageat2024exploring} is another fixed-sampling approach, a variant of ME-Sampling, that modifies the addition of solutions to the archive to account for both their fitness and reproducibility. 
Similarly to ME-Reprod, in our \framework{}, this corresponds to an alternative depth-ordering operator. In the case of ME-Weighted, solutions are compared based on a weighted sum of fitness and reproducibility. 
ME-Weighted was originally introduced with a parametrisation to choose these weights based on user preference on the Performance-Reproducibility Trade-off~\cite{flageat2024exploring}.

\paragraph{AS-Weighted}
AS-Weighted~\cite{flageat2024exploring} was introduced jointly with ME-Weighted and uses the same depth-ordering mechanism, also based on a weighted sum of fitness and reproducibility. The parametrisation to choose the weights from user preferences also applies. 
However, unlike ME-Weighted, AS-Weighted uses all other components from AS. Thus, it uses a depth and an extraction operator that extracts the full content of the archive at every generation.

\paragraph{ME-Low-Spread}
ME-Low-Spread~\cite{mace2023quality} is another fixed-sampling approach, a variant of ME-Sampling, that modifies the addition of solutions to the archive to account for both their fitness and reproducibility. 
Similarly to ME-Reprod, in our \framework{}, this corresponds to an alternative depth-ordering operator. 
In the case of ME-Low-Spread, a new solution can replace an elite in its cell only if it improves both its fitness and reproducibility. 
The other components stay the same as for ME-Sampling and ME.

\paragraph{ME-Delta}
ME-Delta~\cite{flageat2024exploring} can be seen as a generalisation of ME-Low-Spread~\cite{mace2023quality} that enables some more flexible comparisons. ME-Delta relies on the definition by the final user of two thresholds $\delta_f$ and $\delta_r$ respectively for fitness and reproducibility. A solution can replace an existing elite if it either improves the performance of the elite by at least $\delta_f$ or if it improves the performance of the elite without losing more than $\delta_r$ reproducibility or if it improves the reproducibility of the elite by at least $\delta_r$ without loosing more than $\delta_f$ performance. 
This delta comparison allows implementing user preference over the Performance-Reproducibility Trade-off (see Section~\ref{sec:problems}), given in the form of $\delta_f$ and $\delta_r$. It also enables overcoming deceptive reproducibility or fitness traps, in which ME-Low-Spread would get stuck due to its strict criteria. 

\paragraph{AS-Delta}
AS-Delta~\cite{flageat2024exploring} was introduced jointly with ME-Delta and relies on the same delta comparison. However, AS-Delta introduces this comparison with all components from AS. Thus, it uses a depth and an extraction operator that extracts the full content of the archive at every generation.

\paragraph{MOME-X}
MOME-X~\cite{flageat2024exploring} goes one step further than ME-Delta and AS-Delta in accounting for the Performance-Reproducibility Trade-off. MOME-X build on top of existing multi-objective QD works~\cite{pierrot2022multi, janmohamed2023improving} and keeps a Pareto front of fitness and reproducibility trade-offs in each cell of the archive. This allows the user to choose its favourite trade-off a-posteriori, after optimisation. For better estimation, MOME-X relies on the same mechanism as ME-Sampling (i.e. fixed sampling) and samples every solution $32$ times during their first evaluation. 
Thus, MOME-X uses a Multi-Objective ME (MOME)~\cite{pierrot2022multi} container that consists of a grid of $C$ Pareto front. MOME-X also uses selection based on crowding distances within cells from follow-up work~\cite{janmohamed2023improving}. However, MOME-X does not use any extraction or mechanisms that can be assimilated. Thus, in MOME-X, once elites have been evaluated and added to the archive, they cannot be re-evaluated or re-questioned.

\paragraph{ARIA}
The Archive Reproducibility Improvement Algorithm (ARIA)~\cite{grillotti2023don} differs slightly from the approaches described in this section. ARIA was originally designed as an optimization module to improve the reproducibility of an existing archive. However, in the original work, ARIA was also applied from scratch to certain tasks. In this study, we consider this setup to ensure comparability with existing approaches.
Despite this, ARIA’s different motivations mean it was not developed under the same constraints as other UQD approaches. The most significant distinction is its requirement for a significantly higher number of evaluations. In our \framework{}, ARIA employs the same grid container as ME, without depth or extraction mechanisms. However, ARIA utilizes an exceptionally high number of samples per solution ($1024$ to $2048$, depending on the task) and selects only solutions that are at the edge of filled cells in the archive, i.e. neighbouring an empty cell. Finally, ARIA employs a variation operator based on Natural Evolution Strategy~\cite{wierstra2014natural} that leverages a constrained optimization formulation to optimize both fitness and reproducibility simultaneously.


\section{Additional Details on Experimental Setup} \label{app:setup}

We give in Table~\ref{tab:tasks_complete} and Table~\ref{tab:tasks_qdrl_complete} a more detailed experimental setup for our two experimental studies. These tables include the details already presented in the main paper but complement them for a better understanding of the experimental setup. 

\paragraph{Main experimental study}
We give all parameters in Table~\ref{tab:tasks_complete}. 
The details for the Hexapod, Walker, and Ant tasks are adopted from prior UQD research~\cite{flageat2020fast, flageat2023uncertain, grillotti2023don, flageat2024exploring, mace2023quality}. Similarly, for the two Arm tasks, the main parameters align with those used in earlier UQD works~\cite{adaptive, flageat2020fast, flageat2023uncertain}. However, unlike most previous studies, which applied both types of noise simultaneously, we opted to separate them to decorrelate their effects.
We believe this separation offers a clearer understanding of the limitations of existing algorithms. For instance, our main results in Section~\ref{sec:results_algo} demonstrate that Deep-Grid struggles with one type of noise but not the other.

\paragraph{QDRL experimental study}
We give all parameters in Table~\ref{tab:tasks_qdrl_complete}. 
All details are based on previous QDRL works~\cite{nilsson2021policy, tjanaka2022approximating, faldor2023synergizing, mace2023quality} and the original task specifications from OpenAI Gym~\cite{towers2024gymnasium}. Consequently, the uncertainty in these tasks is derived directly from the original OpenAI Gym implementations~\cite{towers2024gymnasium}, without introducing additional elements to increase uncertainty.
For this study, we use neural network controllers with hidden layers of size $(64, 64)$. While this is smaller than the configuration in the original PGA paper~\cite{nilsson2021policy}, it matches the settings used in subsequent works, where it has been shown to maintain PGA’s performance~\cite{tjanaka2022approximating, faldor2023synergizing}.

\input{Tables/tasks_table_complete}

\input{Tables/tasks_qdrl_table_complete}

\begin{figure*}[t!]
  \centering
  \includegraphics[width=0.9\linewidth]{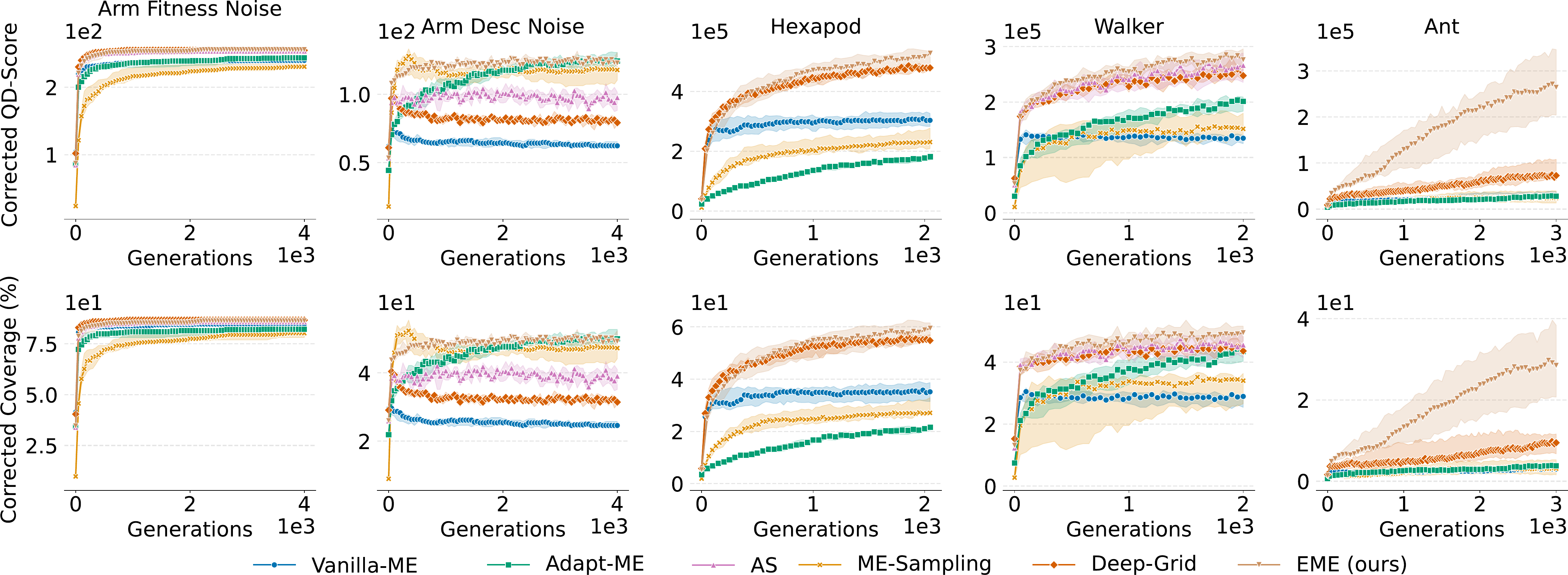}
  \caption{
    Results of our \name{} approach compared to existing UQD approaches tackling the Performance Estimation problem. The plain line indicates the median over $10$ replications, the shaded area corresponds to the quartiles. 
  }
  \label{fig:convergence}
  \Description{Convergence results}
\end{figure*}

\newpage


\section{Additional Results} \label{app:results}

\subsection{Convergence Plots and Run Time for \name{}}

We provide the convergence plots corresponding to the experimental analysis of \name{} in Figure~\ref{fig:convergence}. As detailed in Section~\ref{sec:results_algo}, all algorithms are run with the same evaluation budget per generation (i.e. sampling-size). Therefore, while the x-axis shows the number of generations, it directly translates into the number of evaluations.

\begin{figure}[t!]
  \centering
  \includegraphics[width=\linewidth]{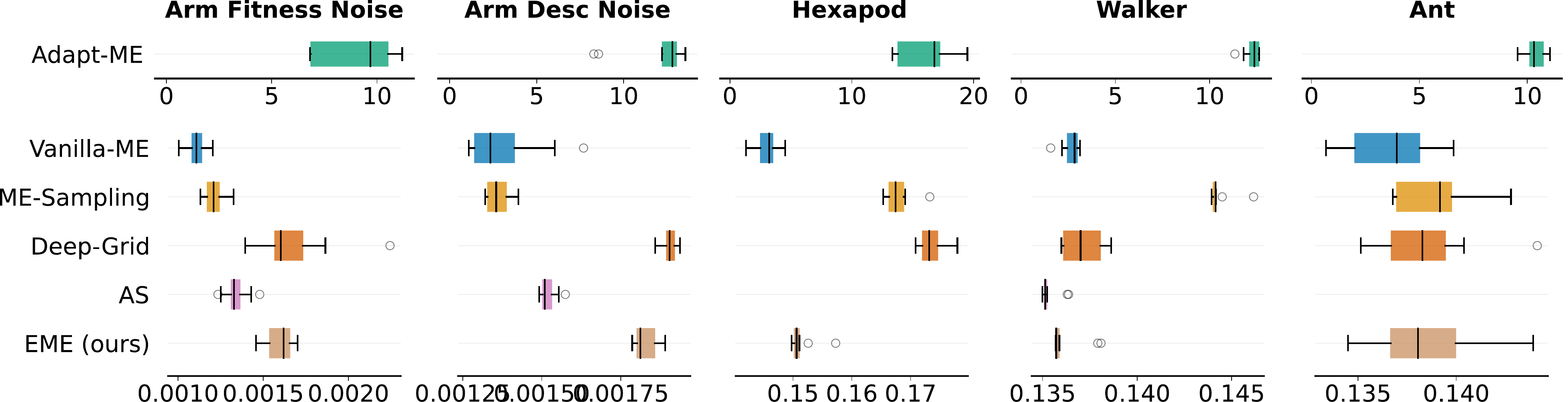}
  \caption{
    Run time in hours taken by all approaches on all tasks, excluding metric computation time. We provide a different axis for Adapt-ME as it is multiple of magnitude slower than other approaches. 
  }
  \vspace{-2mm}
  \label{fig:time}
  \Description{Time Comparison Results}
\end{figure}

We also provide the total run time in hours of our implementations in Figure~\ref{fig:time}. These measures exclude any metric computation time. 
The results show that \name{} does not induce additional time overhead compared to other UQD approaches. 
These results also highlight that Adapt-ME is multiple orders of magnitude slower than other approaches due to its sequential approach.

\subsection{Archives and Convergence Plots for \namepga{}}

\begin{figure}[t!]
  \centering
  \includegraphics[width=0.89\linewidth]{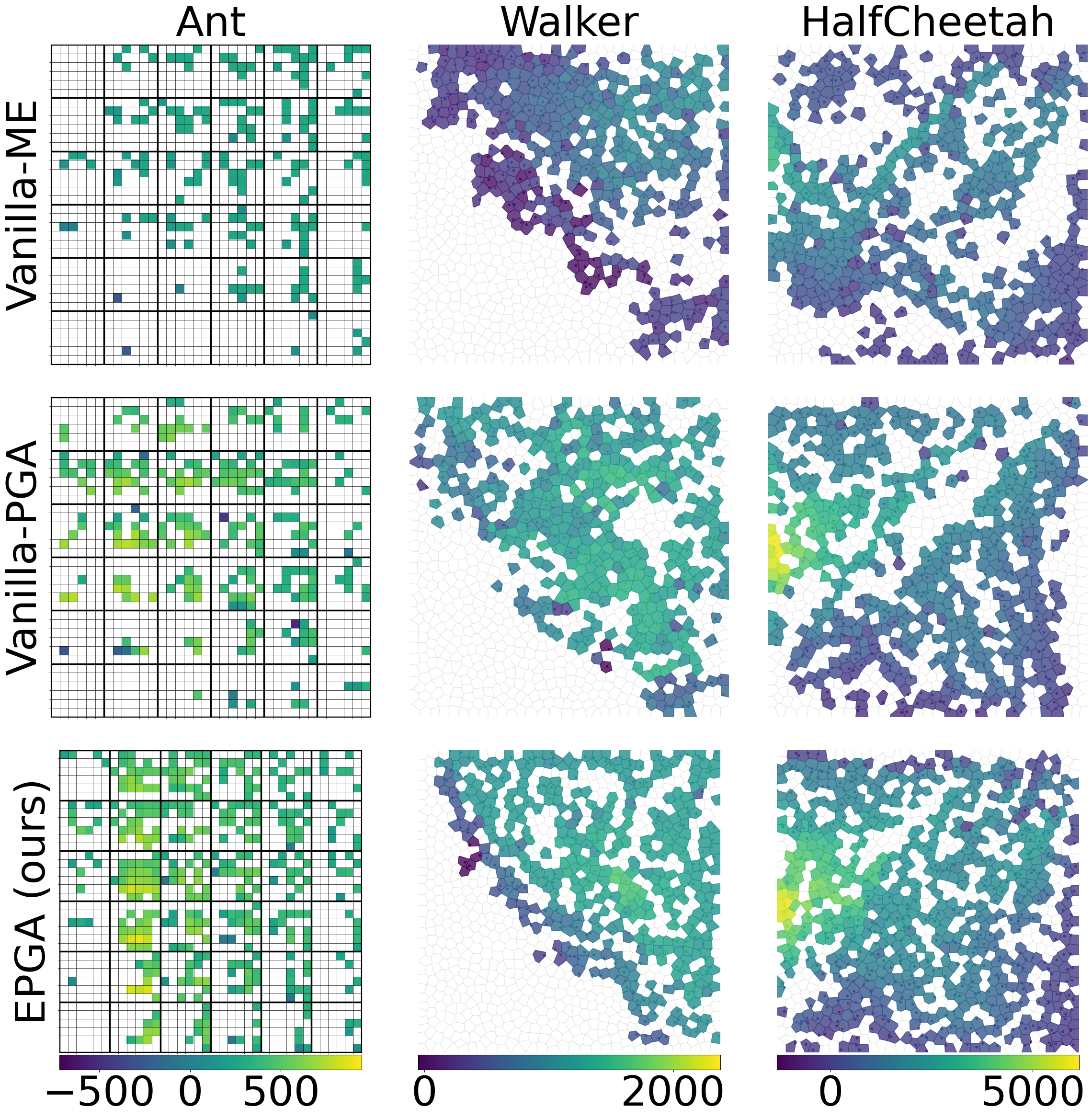}
  \caption{
    Final Corrected Archive of a random replication of each QDRL approach across all tasks. Blank panels indicate undefined approach-task pairs. 
    Each cell corresponds to a solution, with brighter colours indicating higher fitness. The axes represent the descriptor dimensions detailed in Table~\ref{fig:archives_qdrl}.
  }
  \Description{Final archive comparison plots}
  \label{fig:archives_qdrl}
\end{figure}

\begin{figure}[t!]
  \centering
  \includegraphics[width=0.95\linewidth]{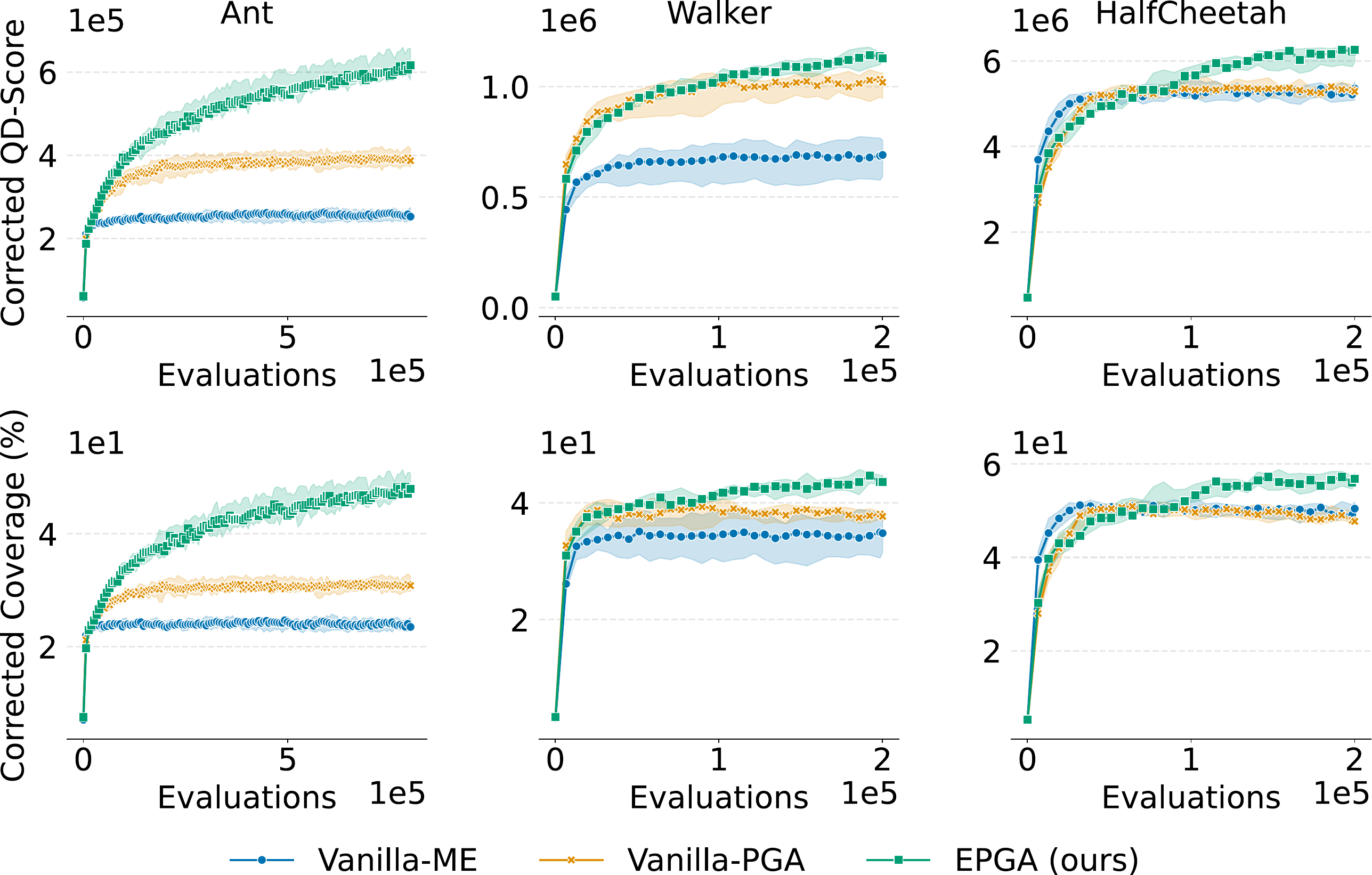}
  \caption{
    Results of our \namepga{}{} approach compared to Vanilla-ME and PGA. The plain line indicates the median over $10$ replications, the shaded area corresponds to the quartiles. 
  }
  \vspace{-2mm}
  \label{fig:qdrl_convergence}
  \Description{Convergence results}
\end{figure}

For completeness, we provide the archives and convergence plots of the experimental analysis of \namepga{} in Figure~\ref{fig:archives_qdrl} and~\ref{fig:qdrl_convergence} respectively.

%% file: Tables/tasks_table_complete.tex
\begin{table}[h]
\centering
\small
  \caption{Detailed task suite used in the main experiments.}
  \begin{tabular}{ c | c c c c }

    & \makecell{\textsc{Arm} \\ \includegraphics[width = 0.06\textwidth]{Figures/arm.png}}
    & \makecell{\textsc{Hexapod} \\ \includegraphics[width = 0.06\textwidth]{Figures/hexapod.png}}
    & \makecell{\textsc{Walker} \\ \includegraphics[width = 0.06\textwidth]{Figures/walker.png}}
    & \makecell{\textsc{Ant} \\ \includegraphics[width = 0.06\textwidth]{Figures/ant.png}}
    \\

    \midrule

    \textsc{Control} 
    & \makecell{Joints \\ position}
    & \makecell{Periodic \\ functions}
    & \multicolumn{2}{c}{\makecell{Neural network}}
    \\

    \textsc{Parameters} 
    & -
    & \makecell{Amplitude \\ and phase}
    & \multicolumn{2}{c}{\makecell{Weights and biases \\ hidden layer of size $8$}}
    \\

    \textsc{\makecell{Observations}} 
    & -
    & -
    & $\mathit{17}$
    & $\mathit{29}$
    \\

    \textsc{\makecell{Action}} 
    & $\mathit{8}$
    & $\mathit{18}$
    & $\mathit{6}$
    & $\mathit{8}$
    \\

    \textsc{\makecell{Control Dims}} 
    & $\mathit{8}$
    & $\mathit{24}$
    & $\mathit{270}$
    & $\mathit{296}$
    \\

    \midrule

    \textsc{Fitness} 
    & \makecell{- Joints \\ position \\ variance}
    & \makecell{Direction \\ error}
    & \makecell{+ Speed \\ - energy \\ + survival}
    & \makecell{- Energy \\ + survival}
    \\

    \midrule
    
    \textsc{Descriptor} 
    & \makecell{End-effector \\ position}
    & \makecell{Final \\ position}
    & \makecell{Feet \\ contact}
    & \makecell{Final \\ position}
    \\

    \textsc{Dims} 
    & $\mathit{2}$
    & $\mathit{2}$
    & $\mathit{2}$
    & $\mathit{2}$
    \\

    \textsc{Nb Cells} 
    & $\mathit{256}$
    & $\mathit{1024}$
    & $\mathit{256}$
    & $\mathit{1024}$
    \\

    \midrule
    
    \textsc{\makecell{Uncertainty}} 
    & \makecell{Fitness OR \\ Descriptor}
    & \makecell{Fitness and \\ Descriptor}
    & \multicolumn{2}{c}{\makecell{Initial joints' position \\ and velocity}}
    \\

    \textsc{Structure} 
    & \makecell{$\mathcal{N}(0, 0.1)$ \\ OR \\ $\mathcal{N}(0, 0.01)$}
    & \makecell{$\mathcal{N}(0, 0.05)$ \\ and \\ $\mathcal{N}(0, 0.05)$}
    & \makecell{$\mathcal{U}(-0.05,$ \\ $0.05)$}
    & \makecell{$\mathcal{U}(-0.1,$ \\ $ 0.1)$}
    \\
  
  \end{tabular}
  \label{tab:tasks_complete}
\end{table}

%% file: Tables/tasks_qdrl_table_complete.tex
\begin{table}[h]
\centering
\small
  \caption{Detailed task suite used in QDRL experiments.}
  \begin{tabular}{ c | c c c }

    & \makecell{\textsc{Walker} \\ \includegraphics[width = 0.06\textwidth]{Figures/walker.png}}
    & \makecell{\textsc{Cheetah} \\ \includegraphics[width = 0.06\textwidth]{Figures/cheetah.png}}
    & \makecell{\textsc{Ant} \\ \includegraphics[width = 0.06\textwidth]{Figures/ant.png}}
    \\

    \midrule

    \textsc{Control} 
    & \multicolumn{3}{c}{Neural network}
    \\

    \textsc{Hidden Layers} 
    & \multicolumn{3}{c}{$2$ layers of size $64$ and $64$}
    \\

    \textsc{\makecell{Observations}} 
    & $\mathit{17}$
    & $\mathit{18}$
    & $\mathit{29}$
    \\

    \textsc{\makecell{Action}} 
    & $\mathit{6}$
    & $\mathit{6}$
    & $\mathit{8}$
    \\
    
    \textsc{\makecell{Control Dims}} 
    & $\mathit{5702}$
    & $\mathit{5766}$
    & $\mathit{6600}$
    \\

    \midrule

    \textsc{Nb Timesteps} 
    & $1000$
    & $1000$
    & $1000$
    \\

    \midrule

    \textsc{Fitness} 
    & \multicolumn{3}{c}{+ speed - energy + survival}
    \\

    \midrule
    
    \textsc{Descriptor} 
    & \multicolumn{3}{c}{Feet contact with the ground}
    \\

    \textsc{Dims} 
    & $\mathit{2}$
    & $\mathit{2}$
    & $\mathit{4}$
    \\

    \textsc{Nb Cells} 
    & $\mathit{1024}$
    & $\mathit{1024}$
    & $\mathit{1024}$
    \\

    \midrule
    
    \textsc{Uncertainty} 
    & \multicolumn{3}{c}{Initial joints' position and velocity}
    \\

    \textsc{Structure} 
    & \makecell{$\mathcal{U}(-0.05, 0.05)$}
    & \makecell{$\mathcal{U}(-0.1, 0.1)$}
    & \makecell{$\mathcal{U}(-0.1, 0.1)$}
    \\
  
  \end{tabular}
  \label{tab:tasks_qdrl_complete}
\end{table}